\documentclass{article}

\usepackage{arxiv}

\usepackage[utf8]{inputenc} 
\usepackage[T1]{fontenc}    
\usepackage{hyperref}       
\usepackage{url}            
\usepackage{booktabs}       
\usepackage{amsfonts}       
\usepackage{nicefrac}       
\usepackage{microtype}      
\usepackage{lipsum}
\usepackage{graphicx}
\usepackage{natbib}
\usepackage{pifont}
\usepackage{bm}
\usepackage{amsmath}
\usepackage{multirow}
\graphicspath{ {./images/} }

\title{Vision Language Model for Interpretable and Fine-grained Detection of Safety Compliance in Diverse Workplaces}

\author{
 Zhiling Chen \\
  School of Mechanical, Aerospace, and Manufacturing Engineering\\
  University of Connecticut\\
  \texttt{Zhiling.chen@uconn.edu} \\
   \And
 Hanning Chen \\
  Department of Computer Science\\
  University of California Irvine\\
  \texttt{hanningc@uci.edu} \\
  \And
 Mohsen Imani \\
  Department of Computer Science\\
  University of California Irvine\\
  \texttt{m.imani@uci.edu} \\
  \And
   Ruimin Chen \\
  School of Mechanical, Aerospace, and Manufacturing Engineering\\
  University of Connecticut\\
  \texttt{ruimin.chen@uconn.edu} \\
  \And
  Farhad Imani \\
  School of Mechanical, Aerospace, and Manufacturing Engineering\\
  University of Connecticut\\
  \texttt{farhad.imani@uconn.edu} \\
}

\begin{document}
\maketitle
\begin{abstract}
Workplace accidents due to personal protective equipment (PPE) non-compliance raise serious safety concerns and lead to legal liabilities, financial penalties, and reputational damage. While object detection models have shown the capability to address this issue by identifying safety items, most existing models, such as YOLO, Faster R-CNN, and SSD, are limited in verifying the fine-grained attributes of PPE across diverse workplace scenarios.
Vision language models (VLMs) are gaining traction for detection tasks by leveraging the synergy between visual and textual information, offering a promising solution to traditional object detection limitations in PPE recognition. Nonetheless, VLMs face challenges in consistently verifying PPE attributes due to the complexity and variability of workplace environments, requiring them to interpret context-specific language and visual cues simultaneously.
We introduce Clip2Safety, an interpretable detection framework for diverse workplace safety compliance, which comprises four main modules: scene recognition, the visual prompt, safety items detection, and fine-grained verification. The scene recognition identifies the current scenario to determine the necessary safety gear. The visual prompt formulates the specific visual prompts needed for the detection process. The safety items detection identifies whether the required safety gear is being worn according to the specified scenario. Lastly, the fine-grained verification assesses whether the worn safety equipment meets the fine-grained attribute requirements. We conduct real-world case studies across six different scenarios. The results show that Clip2Safety not only demonstrates an accuracy improvement over state-of-the-art question-answering based VLMs but also achieves inference times two hundred times faster.
\end{abstract}


\section{Introduction}
Workplace safety remains a critical concern across various industries, including construction, manufacturing, and healthcare, leading to significant efforts in safety training programs, the implementation of protective equipment, and the enforcement of strict safety regulations to reduce accidents and injuries \citep{margaret2015national, albert2021designing}. Although these measures have improved safety awareness and reduced overall incidents, serious workplace accidents and worker injuries continue to persist \citep{bls2022}.
According to the U.S. Bureau of Labor Statistics, the fatal work injury rate for all workers was 3.7 per 100,000 full-time equivalent workers in 2022, reflecting an increase from 3.6 in 2021 and 3.4 in 2020 \citep{bls2024}. Effective accident prevention in high-risk workplaces typically involves several critical measures, including hazard screening, providing appropriate personal protective equipment (PPE), maintaining on-the-job vigilance, and promoting safety awareness and education \citep{Foulis2021}. Among the accident prevention approaches, the provision and proper use of PPE, such as hard hats, safety goggles, gloves, and high-visibility vests stand out as particularly vital \citep{osha2024}. The U.S. Occupational Safety and Health Administration (OSHA) and similar agencies in other countries mandate that all personnel working in close proximity to site hazards wear proper PPE to minimize the risk of injury. In fact, OSHA reports that the proper use of PPE can prevent nearly 40\% of occupational injuries and diseases, while 15\% of injuries resulting in total disability are caused by the failure to wear proper PPE. 

Traditionally, safety inspections have been performed manually, such as supervisors conducting walk-throughs on construction sites to check for hard hats and safety harnesses, or factory managers inspecting assembly lines to ensure workers wearing gloves and protective eyewear. However, such an approach is labor-intensive, time-consuming, and susceptible to human error. Consequently, there has been a progressive shift towards utilizing sensor-based and vision-based systems to enhance the accuracy and efficiency of safety inspections. Sensor-based systems utilize a variety of technologies to monitor PPE usage in real time. For example, radio frequency identification (RFID) in conjunction with Zigbee technologies is commonly employed to detect the presence of PPE on workers and send compliance reports to central units \citep{barro2012real}. Vision-based systems, leveraging advancements in computer vision and machine learning, analyze images captured by cameras to automatically detect PPE. Techniques such as deep learning algorithms can identify PPE, including hard hats, safety goggles, and high-visibility vests by comparing images to predefined models \citep{nath2020deep, abouelyazid2022yolov4}.

\begin{figure}[]
    \centering
    \includegraphics[width = 0.98\textwidth]{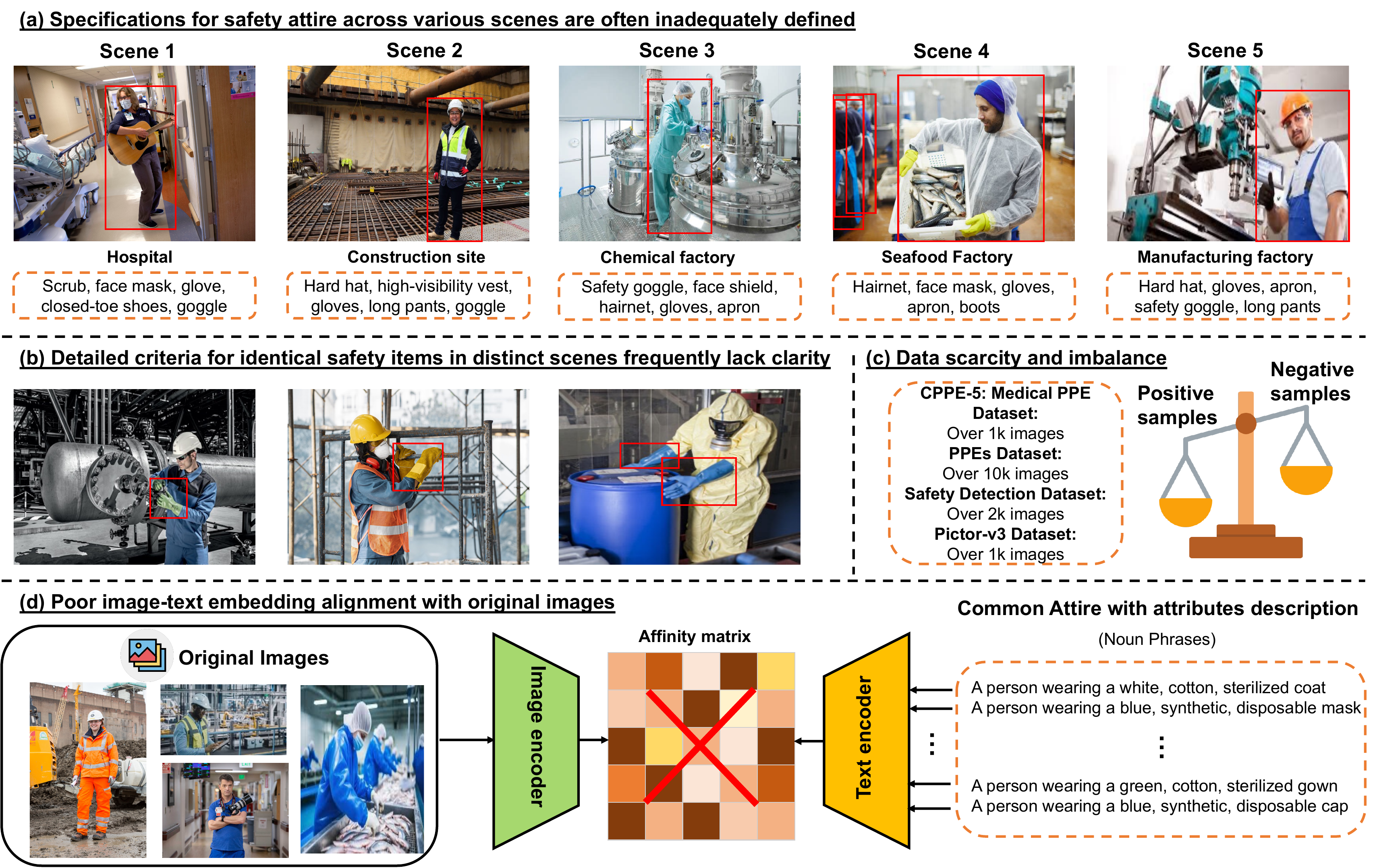}
    \caption{(a) Inadequately defined specifications for safety attire across various scenes. (b) Detailed criteria for fine-grained attributes of different safety attires vary across various scenes (c) Individuals with no safety items only take up a tiny portion of the total samples. (d) Poor image-text embedding when directly using VLMs on original images.}
    \label{intro}
\end{figure}

However, safety requirements vary significantly across diverse and complex work environments, as illustrated in Figure \ref{intro} (a). For example, construction sites mandate hard hats, high-visibility vests, gloves, long pants, and goggles, while chemical factories require safety goggles, face shields, hairnets, gloves, and aprons. Even when similar items are required, their specific attributes can differ; for instance, construction sites demand heavy-duty gloves for mechanical protection, whereas chemical plants require latex gloves resistant to chemical exposure \citep{osha2023}. 
As illustrated in Figure \ref{intro} (b), current models often struggle to adequately address variability in safety equipment and their nuanced attributes across different contexts. This challenge primarily arises because these models are typically trained on limited datasets that focus on specific types of safety gear.  Consequently, models developed for one industry, such as manufacturing, may fail to accurately recognize PPE requirements in other settings, where the types and usage of PPE can differ significantly. 
This can lead to inaccurate results and potentially dangerous oversights. 
Despite efforts to address these issues, many models remain constrained by their reliance on narrow datasets, lacking the generalizability needed for diverse scenarios and tasks. Additionally, sensor-based systems may not capture the full range of environmental variables or worker behaviors, leading to data gaps that compromise safety assessments. Similarly, vision-based systems can suffer from biases in training data, often failing to account for all variations in PPE usage or workplace conditions. 
A recent study \citep{mohona2024yolov8} demonstrates progress in detecting general PPE worn on construction sites; however, it lacks the capability to differentiate between specific scenarios and assess the fine-grained attributes of safety equipment. 
This highlights the critical need for advanced models that can adapt to the specific safety requirements of different industries to enhance overall workplace safety. Furthermore, as highlighted by \cite{johnson2019survey} and shown in Figure \ref{intro} (c), relying solely on sensors or other automated machines for detection can result in insufficient and imbalanced data.

Current deep learning-based object detection models follow either one-stage or two-stage detectors to tackle the challenges \citep{lu2020mimicdet}. The former simultaneously performs classification and regional proposal to obtain results, while two-stage detectors perform classification and regional proposal sequentially. However, both approaches struggle to effectively address the issues of scene and task diversity, as well as data scarcity and imbalance. 
As transfer learning continues to advance, there is a growing emphasis on improving generalization capabilities and addressing data-related challenges \citep{torrey2010transfer}. 
Techniques such as vision language models (VLMs) are being explored to enhance model performance in diverse scenarios by leveraging pre-trained models on large datasets and adapting them to specific tasks with limited data \citep{bordes2024introduction}.
Moreover, integrating additional context-awareness and reasoning capabilities into vision language models helps ensure that detected objects meet the specific standards required for particular tasks.
For instance, in a chemical plant, a system may not only detect that a worker is wearing gloves but also verify that the gloves meet the necessary chemical resistance standards.

However, as shown in Figure \ref{intro} (d), complex environments lead to poor alignment between image and text embeddings, posing new challenges for the accurate detection of safety items. To address this issue, we leverage the object detection model to isolate the individuals or items, thereby improving the alignment of visual and textual embeddings.
Based on this improved alignment, visual language models form the backbone for matching image and text embeddings in object detection, yet often lack the reasoning capabilities needed to identify correct objects in safety detection.
Our framework draws inspiration from human reasoning processes, which are naturally multi-staged. Initially, humans identify potential instances of objects and extract their visual features. Simultaneously, they parse the task at hand, discerning the necessary visual or functional attributes of items that workers need to wear in specific scenarios. Subsequently, they evaluate the instances based on task-relevant attributes to determine whether the individuals or items meet safety requirements. Humans do not merely judge whether an item meets a standard; instead, they engage in a step-by-step reasoning process, using the attributes required by the specific task scenario as clues.


In particular, we propose a zero-shot learning framework for fine-grained safety violation detection across various workplaces. Our framework adopts visual language models and large language models to align the information between workspace image data and text information and consists of two steps. First, we implement an object detection model, such as YOLO, to identify and extract bounding boxes of individuals and their respective safety gear within the image. This step ensures that each person and safety equipment are isolated for further analysis. Second, we employ a vision language model to compare the cropped images with textual descriptions, such as "a person wearing [item]" for initial detection and "a [feature] [item]" for attribute verification. This step ensures that the detected safety equipment not only matches the required items but also meets the specified attributes essential for safety compliance. The framework has been implemented on real-world safety compliance datasets from six distinct environments, each with limited data, to evaluate its capability across diverse workplaces.
The contribution of this work is as follows:

\begin{itemize}
    \item Instead of training a multi-modal model from scratch, we propose Clip2Safety, a VLM-based, multi-module, and dynamic safety compliance framework that leverages the semantic information from vision-language pretraining and the capabilities of object detection models to support a calibrated vision-text embedding space.

    \item We design a scene recognition module to identify scenarios and determine the necessary safety gear, ensuring adequate matching between scene-specific requirements and visual prompts.
    
    \item We introduce a visual prompt module that formulates visual prompts needed for the detection process, enhancing the model's ability to adapt to varying safety compliance requirements.

    \item Our framework not only demonstrates significant improvements in accuracy over state-of-the-art question-answering based VLMs but also achieves inference times that are two orders of magnitude less, showing both high efficiency and effectiveness in dynamic safety compliance scenarios.
    
\end{itemize}

The remainder of this paper is organized as follows: 
Section~\ref{background} reviews PPE detection, vision language model.
Section~\ref{methodology} introduces the Clip2Safety framework for model implementation. Section~\ref{Experiment} discusses the dataset and experiments. Finally, the conclusion of this proposed study is presented in Section~\ref{Conclusion}.

\section{Research Background} \label{background}

\subsection{Personal Protective Equipment Detection}

The detection of PPE has been significantly advanced through the application of deep learning and computer vision technologies. 
Early methods for PPE detection primarily relied on sensor-based systems, such as RFID technology \citep{barro2012real}.
For example, one approach involved equipping each PPE component with RFID tags and using scanners at job site entrances to ensure workers were wearing the appropriate gear \citep{kelm2013mobile}. 
Another method investigated the use of short-range transponders embedded in PPE items, with a wireless system verifying compliance with safety regulations \citep{naticchia2013monitoring}.
Similarly, \citep{barro2012real} investigated a local area network to monitor RFID tags on PPE continuously, ensuring compliance throughout the workday. 
Additionally, GPS technology has also been utilized to enhance PPE detection. For instance, \citep{zhang2015global} employed GPS devices attached to workers' safety helmets to provide supplementary safety monitoring on construction sites. In a more recent advancement, \citep{pisu2024enhancing} introduced an operator area network system that leverages Machine Learning and RSSI analysis to detect improper PPE usage robustly. This system uses an SVM model with a customizable post-processing algorithm, reducing false positives by 80\% and detecting issues within seven seconds. Furthermore, \citep{yang2020automated} designed an automated PPE-tool pairing detection system based on a wireless sensor network, which effectively monitors the wearability status of PPE. 
However, despite their effectiveness in certain scenarios, these sensor-based approaches were limited by their dependency on physical sensors attached to PPE items and necessitated significant investment in purchasing, installing, and maintaining complex sensor networks, which could hinder their practical implementation \citep{nath2020deep}. 

With the advancement of computer vision technology, vision-based systems have revolutionized PPE detection due to their lower cost, simpler configuration, and broader application scenarios compared to sensor-based technologies. These systems utilize cameras to capture real-time images and videos of workers, which are then analyzed using computer vision techniques. The extensive deployment of cameras, combined with significant progress in computer vision technology, has established a foundation for effective PPE detection. 
For instance, \citep{wu2018intelligent} proposed a color-based hybrid descriptor that combines local binary patterns, human movement invariants, and color histograms to extract features of hardhats in various colors, which are then classified using a hierarchical support vector machine (HSV). \citep{li2018automatic} developed a method for detecting workers utilizing the ViBe and the C4 framework, leveraging the HSV color space for hardhat classification. \citep{mneymneh2019vision} introduced a framework that first detects moving objects using a standard deviation matrix and then classifies them with an aggregate channel feature-based object detector. This approach integrates a histogram of oriented gradients-based cascade object detectors to identify hardhats in the upper regions of detected personnel, which are subsequently processed by a color-based classification component. Despite their effectiveness, these multi-stage methods heavily depend on hand-crafted features and encounter difficulties in complex scenes with varying conditions, different viewpoints and occlusions.

Thus, convolutional neural networks (CNNs) \citep{lecun2015deep} have become the backbone of the systems due to their robust image recognition capabilities. Recent studies employing CNNs for object detection have primarily utilized faster region-based CNNs (R-CNNs) \citep{ren2015faster} and you only look once (YOLO) \citep{redmon2016you}. These models facilitate the recognition of target objects, such as persons and helmets. When trained with sufficiently large datasets, they exhibit robust and improved performance, significantly enhancing the reliability of vision-based PPE detection systems.
Faster R-CNN incorporates a region proposal network to improve detection speed and accuracy. For example, \citep{saudi2020image} leverages Faster R-CNN to check workers’ safety conditions based on PPE compliance. Also, \citep{chen2020detection} introduced retinex image enhancement to improve image quality for the outdoor complex scenes in substations based on Faster R-CNN. 
On the other hand, YOLO is known for its speed and efficiency making it suitable for real-time applications. YOLO divides the image into a grid and predicts bounding boxes and class probabilities directly from full images in one evaluation. Many researchers have utilized variants of YOLO for PPE detection. For instance, \citep{wu2019helmet} utilizes the advantage of Densenet in model parameters and technical cost to replace the backbone of the YOLO V3 network for feature extraction, forming a YOLO-Densebackbone convolutional neural network to improve helmet detection and ensure safe construction. \citep{chen2023lightweight} introduced a lightweight WHU detection model called WHU-YOLO, which enhances YOLOv5s with a Ghost module and Bi-FPN. Additionally, \citep{benyang2020safety} used a multi-scale training strategy to enhance the adaptability of the YOLO V4 from different scales of detection.
To enhance detection accuracy, researchers have increasingly integrated object detection models with classification models, resulting in more precise and reliable safety assessments. For instance, \citep{lee2023verification} proposed a method that combines the strengths of YOLO and CNN-based approaches. In this two-stage process, YOLO is first employed to detect whether a helmet is being worn. Subsequently, a CNN-based classification model is used to distinguish between helmets, heads, and hats, allowing for a more nuanced evaluation of PPE compliance.

Despite these advancements, most research has focused on specific, high-supervision environments, such as construction sites, industrial plants, and nuclear facilities, where clear safety protocols and structured activity schedules are in place \citep{chen2020vision, onal2021object}. 
In dynamic and less supervised environments, such as manufacturing laboratories or outdoor construction sites, the requirements for PPE usage can vary significantly depending on the time of day, specific tasks being performed, and changing environmental conditions. Consequently, there is a pressing need for interpretable detection methods that can adapt to diverse scenarios and provide insights into the specific safety requirements of each context. This necessitates a system capable of not only detecting PPE but also understanding the contextual factors influencing its usage, thereby ensuring comprehensive safety compliance across different settings \citep{fang2018detecting, hung2021unsafe}.

\subsection{Vision Language Model}

Recent years have witnessed substantial success in extending pre-trained vision language models to support new applications. Among the most successful efforts, models like Flamingo \citep{alayrac2022flamingo}, OpenAI CLIP \citep{radford2021learning}, and OpenCLIP \citep{cherti2023reproducible} have exhibited impressive performance in handling image-text matching, owing to their semantic knowledge and understanding of content that spans both modalities. These models have been applied successfully in downstream applications such as object detection \citep{shi2022proposalclip}, image captioning \citep{mokady2021clipcap}, action recognition \citep{wang2021actionclip}, task-oriented object detection \citep{chen2024taskclip}, anomaly segmentation \citep{jeong2023winclip}, semantic segmentation \citep{liang2023open}, and dense prediction \citep{zhou2023zegclip}. However, existing CLIP-based algorithms primarily focus on matching image patches with nouns in text, which poses challenges in understanding different people and objects within the images. Therefore, additional modules are needed to facilitate the matching between the visual attributes of image patches and the adjective phrases describing different people and items.

To enhance the alignment between images and text, a recent application of VLMs for PPE detection introduced a three-step zero-shot learning-based monitoring method \citep{gil2024zero}. First, it detects workers on-site from images and crops body parts using human-body key points. Next, the cropped body images are described with image captioning. Finally, the generated text is compared to prompts describing PPE-wearing body parts, determining safety based on cosine similarity. However, this method still has limitations in accurately distinguishing between different types of PPE and their specific features in diverse environments.

Alongside these advancements, significant progress has been made in question-answering based VLMs, which is usually called visual question answering models (VQA). \citep{bulian2022tomayto} pioneered the creation of the first free-form and open-ended VQA dataset, where human workers were asked to create questions that a smart robot might not answer and then collect human answers per question. The follow-up work, VQAv2 \citep{goyal2017making}, enhanced the previous dataset to reduce statistical biases in the answer distribution. Instead of relying on human workers, the GQA dataset \citep{hudson2019gqa} employed scene graph structures from visual genome \citep{krishna2017visual} to generate question-answer pairs. These graph structures allowed the authors to balance the answer distributions for each question, reducing the dependency on answer statistics. Both datasets have been widely used for various VQA tasks \citep{alayrac2022flamingo, dai2023instructblipgeneralpurposevisionlanguagemodels}. Due to the question-answer format of VQA models, it aligns more naturally with the process of inspecting safety equipment. \citep{ding2022safety} formulated a "rule-question" transformation and annotation system, turning safety detection into a visual question answering task, and leverages the strengths of VQA models to enhance the accuracy and efficiency of PPE compliance checks.

Despite these advancements, a notable tradeoff exists between achieving high-performance VQA models and maintaining acceptable inference times. High-performing VQA models typically require significant computational resources and time, leading to longer inference times. However, real-time safety detection applications impose strict requirements on inference times to ensure timely and effective responses. Thus, making a new model for such applications necessitates a careful balance between performance and inference speed.

\section{Research Methodology} \label{methodology}
The proposed Clip2Safety, an interpretable and fine-grained detection framework for diverse workplace safety compliance, comprises four main modules: scene recognition, the visual prompt, safety items detection, and fine-grained verification.

\begin{figure}[]
    \centering
    \includegraphics[width = 0.98\textwidth]{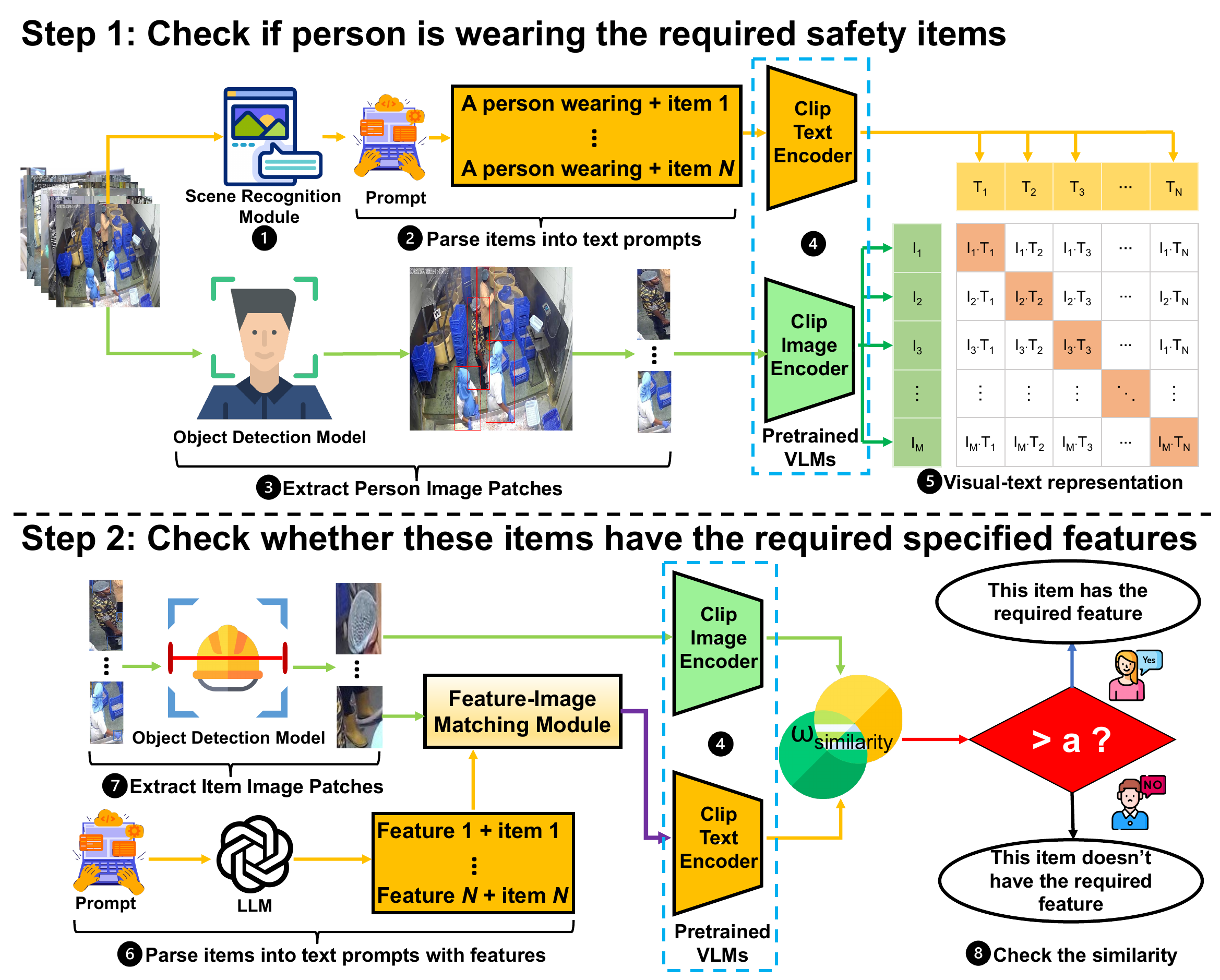}
    \caption{Model Architecture: Step 1: Detect if the person is wearing the required safety items using scene recognition and object detection, paired with VLMs for verification. Step 2: Verify that the detected safety items meet specific attribute requirements by comparing image patches with generated text prompts using a feature-image matching module.}
    \label{model}
\end{figure}

\subsection{Scene Recognition Module}
To address the challenges posed by the diversity of real-world scenes and the corresponding variety of safety gear requirements shown in Fig.\ref{intro}(a), we developed a scene recognition module to identify scenes in images. As illustrated in Fig.\ref{model} \ding{182}, this module generates high-quality image captions that provide concise summaries of the visual content. However, this abstraction process can result in the loss of specific visual details, potentially affecting our framework's performance. To tackle this issue, we conducted a comprehensive survey of existing image captioning models, prioritizing those capable of generating more accurate scene descriptions while also considering our computational resource constraints. Ultimately, we selected Salesforce's BLIP2-OPT-2.7B model as our image captioning model due to its advanced multi-modal capabilities, which integrate both vision and language understanding, allowing it to generate more nuanced and context-aware descriptions.

To further improve the performance of the scene recognition module while adhering to computational resource constraints, we employed Low-Rank Adaptation (LoRA) \citep{hu2021lora} to provide accurate scene recognition by fine-tuning a minimal number of weights. 
For this purpose, we used the text prompt "in the [scene]" paired with the respective image as the textual input. 
LoRA works by introducing small additive weights into existing linear weight matrices. These weights retain the original weight matrix's dimensions and provide a parallel trainable pathway for the incoming feature map when multiplied. The core idea is that during adaptation the updates to the weight matrices have low intrinsic ranks.

The modification involves adding a rank-constrained product of matrices $\bm{A}$ and $\bm{B}$. For a weight matrix $ \bm{W} \in \mathbb{R}^{d \times k} $, the update is represented as $ \bm{W}' = \bm{W} + \bm{B}\bm{A} $, where $ \bm{B} \in \mathbb{R}^{d \times r} $ and $ \bm{A} \in \mathbb{R}^{r \times k} $, and $ r \ll \min(d,k) $. Here, $ d $ and $ k $ are the dimensions of $ \bm{W} $, and $ r $ is the rank of the adapters. During fine-tuning, $\bm{W}$ remains unchanged, while the weights of $\bm{A}$ and $\bm{B}$ are updated. Following the initialization method in \cite{hu2021lora}, $\bm{B}$ is initialized to zeros, and $\bm{A}$ with small random values from a Gaussian distribution. The updated matrix $\bm{W}'$ is given by:

\begin{equation}
\bm{W}' = \bm{W} + \frac{\alpha}{r} \bm{B}\bm{A}.
\end{equation}

In this equation, $ \alpha $ is a scaling parameter that adjusts the influence of the new weights on the original weights. 
We implement this approach to the two weight matrices $ \bm{W}_{q}, \bm{W}_{k}$, corresponding to the query, key, and output matrices in the multi-head self-attention module of the transformer architecture. The fine-tuning process aims to minimize the negative log-likelihood loss $\mathcal{L}$, defined as:
\begin{equation}
\mathcal{L} = -\sum_{i=1}^{\bm{K}} \log P(C_{i} | \bm{I}_{i}; \boldsymbol{\theta})
\end{equation}
where $P(C_{i} | I_{i}; \boldsymbol{\theta})$ is the probability of generating the correct caption $C_{i}$ given the data sample $\bm{I}_{i}$ and model parameters $\boldsymbol{\theta}$, includes the adapted weights $\bm{W}'$, and $\bm{K}$ is the total number of data samples.

\subsection{Visual Prompt Module} \label{Visual Prompt Module}

\begin{figure}[]
    \centering
    \includegraphics[width = 0.98\textwidth]{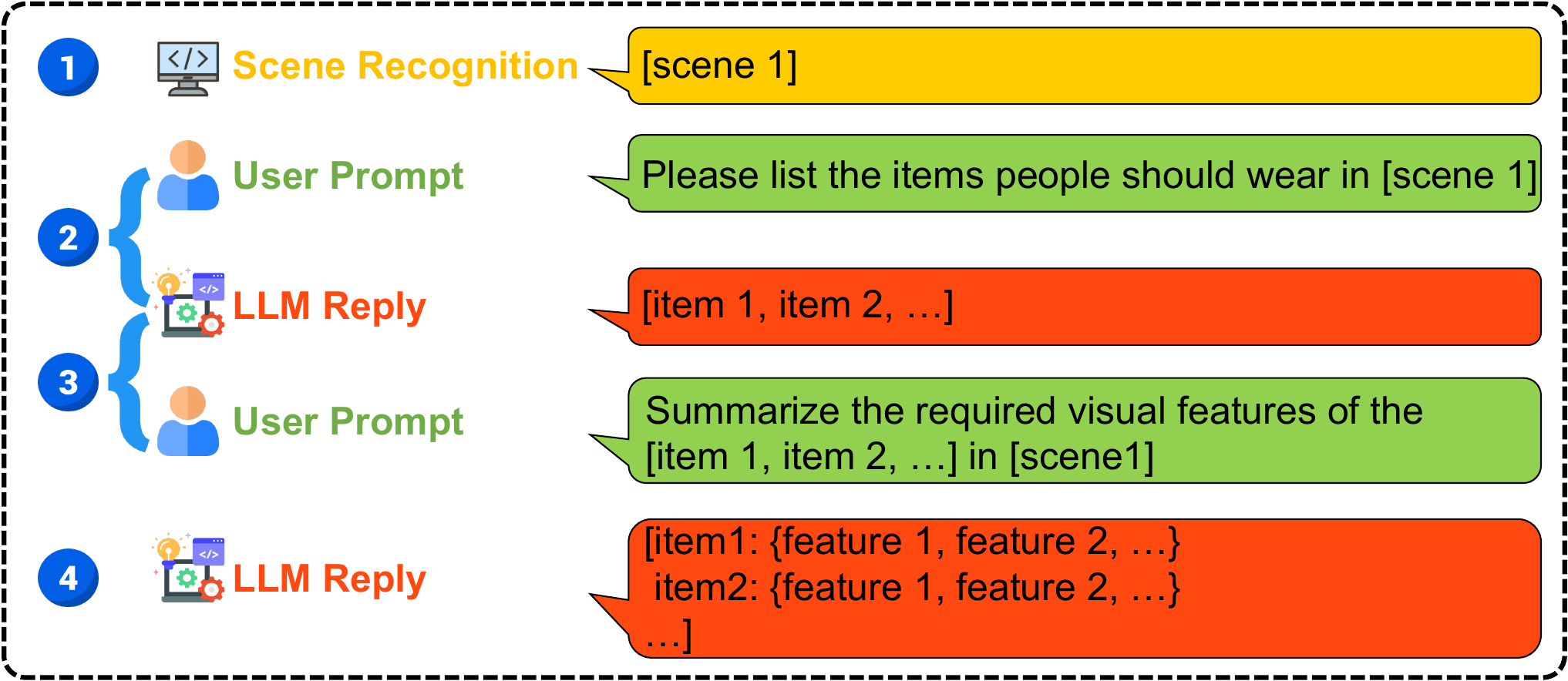}
    \caption{Clip2Safety Visual Prompt Module: Beginning with scene recognition to identify the environment, user prompts are then issued to a large language model to retrieve the necessary safety items and their specific visual features relevant to the recognized scene.}
    \label{prompt}
\end{figure}


In the visual prompt module, we address the diversity of safety requirements by employing a large language model to generate the necessary safety items based on the scene, as depicted in Fig.\ref{model} \ding{183}. To bridge the semantic and reasoning gap between the fine-grained requirements and the safety items, we also employ the LLM to generate scene-relevant visual attributes, as illustrated in Fig.\ref{model} \ding{186}.
In Fig.\ref{prompt}, we provide a detailed description of the prompts utilized to guide the LLM in generating the appropriate safety items and their visual attributes. To illustrate this process, we take a seafood factory as an example. The specific procedure unfolds as follows:

\begin{itemize}
    \item [(1)]
    First, we receive the scene information from the image caption model. Based on our compiled and synthesized dataset, the scenes are as follows: hospital, construction site, chemical factory, seafood factory, and manufacturing zone.
    \item [(2)]
    Then we operate the scene information generated by the image caption model to query the LLM: "List the items people should wear in a seafood factory." The response generated by the LLM encompasses several safety items, such as "hairnet", "face mask", "gloves", "aprons", and "boots".
    \item [(3)] 
    Building upon the LLM's response, we proceed to the second prompt: "Summarize the required visual features of the ["hairnet", "face mask", "gloves", "aprons", "boots"] in a seafood factory." This prompt is designed to guide the LLM in summarizing the specific visual attributes required for the requested safety items, as the same item may have different fine-grained requirements in different scenarios, as is shown in Fig.\ref{intro}(b).
    \item [(4)]
    Finally, we obtained the visual attributes that each safety item people in the seafood factory should wear from LLM responses. For instance, the "boots" should have visual attributes such as "high-visibility color", "Non-slip soles" and "waterproof". In total, we generated three attributes for each item, which belong to color, material, and functionality respectively.
\end{itemize}

\subsection{Safety Items Detection Module} \label{Safety Items Detection Module}

In the safety items detection module, we generate $\bm{N}_{\text{pbbox}}
$ bounding boxes for all the person in the scene, where pbbox denotes person bounding box, as illustrated in Fig.\ref{model} \ding{184}. As mentioned in Fig.\ref{intro} (c), obtaining a multi-scene dataset suitable for the safety compliance task is extremely challenging. Additionally, most single-scene safety detection datasets suffer from severe class imbalance, as only a few workers in the images fail to wear the required items, resulting in the majority being marked as positive samples. To deal with the data scarcity and imbalance problem, we leverage an open vocabulary object detection model, specifically YOLO-World, which is pre-trained on large-scale datasets and with a strong open vocabulary detection capability. Note that, in the basic setting of our framework during inference, the open vocabulary object detection model is only responsible for generating bounding boxes for all the persons in the image. Based on the bounding box coordination, we extract $\bm{N}_{\text{pbbox}}
$ person image patches. After generating $\bm{N}_{\text{pbbox}}
$ person image patches and $\bm{N}_{\text{item}}
$ prompts, we pass them into pre-trained VLMs, such as OpenAI CLIP \citep{radford2021learning}, to generate text and image embeddings as is shown in Fig.\ref{model} \ding{185}. Here we utilize $\bm{L}_{\text{pbbox}}
$ and $\bm{L}_{\text{item}}
$ to represent the lists of person image patches and item prompt texts. The lengths of the lists are $\bm{N}_{\text{pbbox}}
$ and $\bm{N}_{\text{item}}
$, respectively. The computation process can be summarized as follows:
\begin{equation}
\bm{E}_{\text{person}} = \text{CLIP}_{\text{image}}(\bm{L}_{\text{pbbox}})
\end{equation}
\begin{equation}
\bm{E}_{\text{item}} = \text{CLIP}_{\text{text}}(\bm{L}_{\text{item}})
\end{equation}

Suppose the embedding dimension is d, the shape of the generated vision embedding matrix and text embedding matrix will be $\bm{N}_{\text{pbbox}}
 \times$ d and $\bm{N}_{\text{item}} \times$ d. After we get the person vision embedding $\bm{E\textsubscript{\textnormal{person}}}$ and item text embedding $\bm{E\textsubscript{\textnormal{item}}}$, we perform matrix to matrix multiplication between $\bm{E\textsubscript{\textnormal{person}}}$ and $\bm{E\textsubscript{\textnormal{item}}}$ to generate the predicted affinity matrix $\bm{A\textsubscript{\textnormal{predict}}}$, as is shown in Fig.\ref{model} \ding{186}. The computation could be summarized as:
\begin{equation}
\bm{A}_{\text{predict}} = \bm{E}_{\text{person}}^{\top} \times \bm{E}_{\text{item}}
\end{equation}

The shape of the affinity matrix $\bm{A}_{\text{predict}}$ is $\bm{N}_{\text{pbbox}} \times \bm{N}_{\text{item}}$. In traditional CLIP, to perform the zero-shot image classification, we will directly apply the softmax function over the affinity matrix to obtain the prediction. In this step, rather than performing image classification, we need to verify if the person is wearing the required items. For inference, we set a threshold value ($\delta$), and we proceed as follows:
\begin{equation}
\bm{P}_i = 
\begin{cases} 
1 & \text{if } \bm{P}_i \geq \delta, \\
0 & \text{else}.
\end{cases}
\end{equation}

Here, $\bm{P}_i$ represents the prediction of whether the person is wearing the item. To optimize our results, we also adopted an LLM-driven approach, where the affinity matrix is provided to the LLM, allowing it to make the decision for us. As shown in Table \ref{benchmark}, using GPT-4o as the decision-making LLM yielded better results.

\subsection{Fine-grained Verification Module} \label{Fine-grained Verification Module}
In the fine-grained verification module, we leverage the result from the safety items detection module. If the person in the image wears the item, we utilize the same open vocabulary object detection model to generate the bounding boxes of the items and extract the image patches, as is shown in Fig.\ref{model} \ding{188}. As illustrated in Table \ref{attributes}, three distinct types of attributes are generated for each item within the visual prompt module. 

After we generate the prompts for each attribute of each item and the corresponding item's image patch, we pass them into the same pre-trained VLMs to generate text and image embeddings. Here we utilize $\bm{L}\textsubscript{ibbox}$ and $\bm{L}\textsubscript{attribute}$ to represent the lists of item image patches and the attribute's prompt texts, where ibbox means item bounding box. 
The lengths of the lists are $\bm{N}\textsubscript{ibbox}$ and $\bm{N}\textsubscript{attribute}$, respectively. 
The computation process can be summarized as follows:
\begin{equation}
\bm{E}_{\text{item}} = \text{CLIP}_{\text{image}}(\bm{L}_{\text{ibbox}})
\end{equation}
\begin{equation}
\bm{E}_\text{{attribute}} = \text{CLIP}_{\text{text}}(\bm{L}_{\text{attribute}})
\end{equation}
where $\bm{E\textsubscript{\textnormal{item}}}$ and $\bm{E\textsubscript{\textnormal{attribute}}}$ are the item vision embedding matrix and attribute text embedding matrix, respectively. Then we extract the embedding vectors of each item image patch and its corresponding attributes' prompt text. We denote them as  $ \bm{v}_\text{item} $ and $ \bm{v}_\text{attribute} $.
During the feature verification process, we compute the cosine similarity between the item image patch and attribute prompt text embedding vectors to measure their alignment. The cosine similarity $ \bm{\omega}(\bm{v}_\text{item}, \bm{v}_\text{attribute}) $ is calculated as:

\begin{equation}
\bm{\omega}(\bm{v}_\text{item}, \bm{v}_\text{attribute}) = \frac{\bm{v}_\text{item} \cdot \bm{v}_\text{attribute}}{\|\bm{v}_\text{item}\| \|\bm{v}_\text{attribute}\|}
\end{equation}

In this step, we need to check whether the item worn by the person meets the attribute requirements. So we set another threshold value ($\tau$) and we have:
\begin{equation}
\bm{P}_j = 
\begin{cases} 
1 & \text{if } \bm{P}_j \geq \tau, \\
0 & \text{else}.
\end{cases}
\end{equation}

Here $\bm{P}_j$ is the prediction of whether the item worn by the person meets the attribute requirements. We can also leverage the LLM-driven approach to optimize the result. However, this step is different from \ref{Safety Items Detection Module}. We will generate a similarity list for all the items' corresponding attributes, and then we give the list to GPT-4o to make desicion for us.

\subsection{Evaluation}

Ensuring compliance with safety item requirements involves detecting whether a person is wearing the necessary safety items and verifying that these items meet specified attribute standards. We structure this task into three stages: safety items detection, feature verification, and overall evaluation. For each stage, we define different metrics to comprehensively evaluate model performance. 

The first stage, safety items detection, focuses on identifying the presence of required safety items, such as hard hats, safety goggles, and high-visibility vests, on individuals. We choose five required safety items for each environment, generated from LLM, as shown in \ref{Visual Prompt Module}. 
This stage leverages advanced object detection algorithms to accurately pinpoint and label each safety item within the visual input. The detection accuracy is measured by determining the proportion of safety items correctly identified by the model relative to the total number of items that should be present.
To express this more precisely, the total number of individuals is considered, along with the actual number of safety items each person is wearing. The detection accuracy is then calculated by comparing the number of safety items the model correctly identifies with the expected number of safety items across all individuals. 

The second stage, feature verification, involves validating that the detected safety items adhere to specific attribute requirements. For example, it is not only important to detect the presence of gloves but also to ensure that they comply with standards for chemical resistance or thermal protection, depending on the operational context. As detailed in table \ref{attributes}, the attributes are classified into three distinct categories: Directly Observable (DO), Situationally Observable (SO), and Inferentially Observable (IO), with each attribute being assessed individually.
To assess the accuracy of the feature verification for each attribute type, we compare the number of correctly detected attributes within each category (DO, SO, and IO) against the total number of detected items. For Directly Observable attributes, the accuracy is calculated by determining the proportion of correctly identified attributes out of the total items detected. The same approach is applied to Situationally Observable and Inferentially Observable attributes, where the accuracy is measured by comparing the correctly identified attributes in each category to the total number of detected items.

\section{Experiment} \label{Experiment}

\subsection{Dataset}

\begin{figure}[]
    \centering
    \includegraphics[width = 0.98\textwidth]{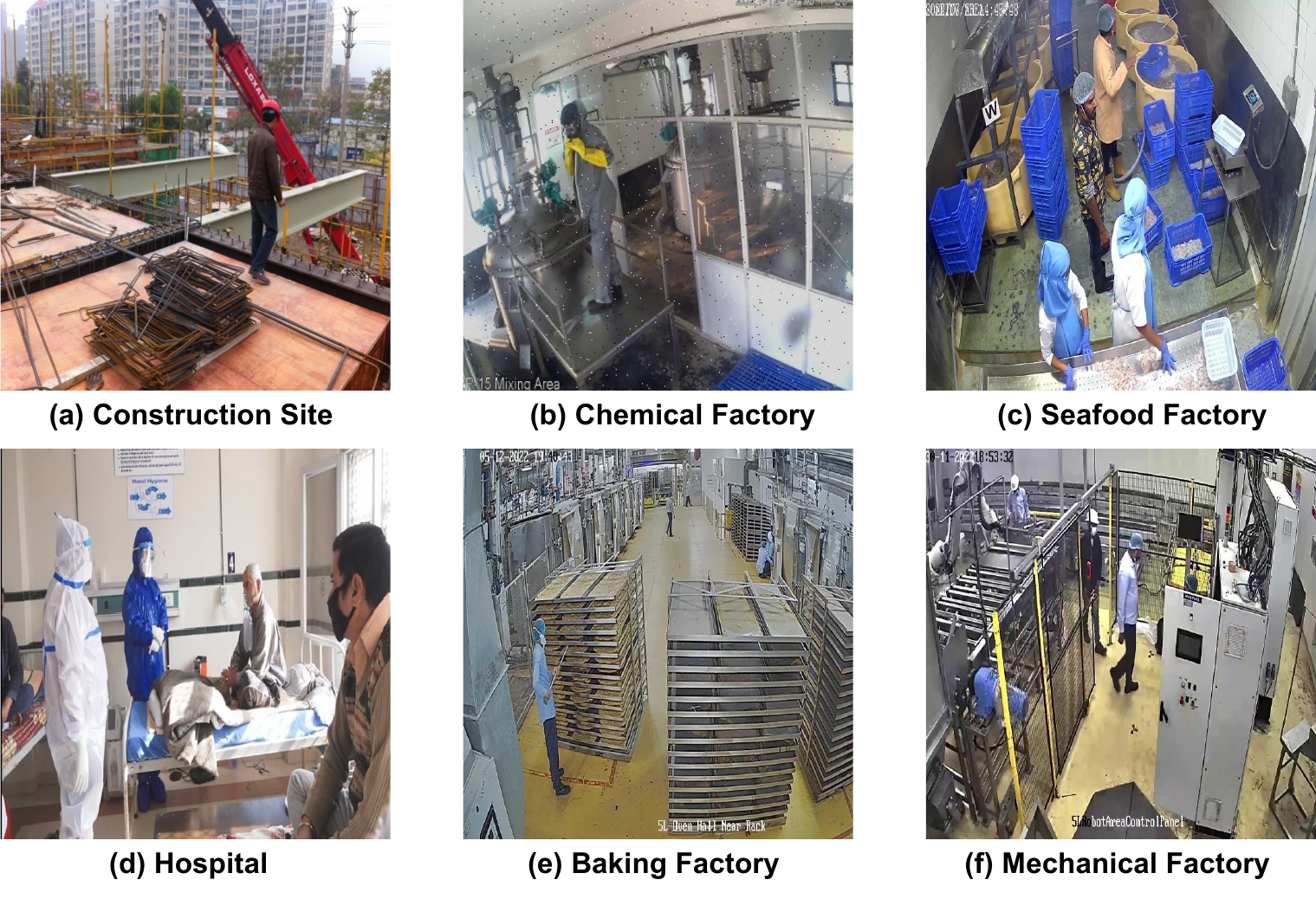}
    \caption{Example images for 6 scenes. (a) Construction site images from Pictor-v3. (b) Chemical Factory images from Safety Detection dataset. (c) Seafood Factory images from PPEs Dataset. (d) Hospital images from CPPE-5 Dataset. (e) Baking Factory images from Safety Detection Dataset. (f) Mechanical Factory images from Safety Detection Dataset.}
    \label{data}
\end{figure}

To meet the diverse scene requirements of our model, we collected and integrated data from four sources due to the lack of a suitable comprehensive public dataset. From these sources, we selected five distinct scenes, and each dataset is described in detail as follows:

\begin{itemize}
    \item \textbf{Safety Detection Dataset  \citep{safety-detection-ge25h_dataset}:} The safety detection dataset, available on Roboflow, targets various safety detection applications. It features images of PPEs in industrial environments such as construction sites and factories. The dataset annotates different types of safety equipment, including hard hats and reflective vests, facilitating effective detection in diverse working conditions.
    \item \textbf{PPEs Dataset  \citep{ppes-kaxsi_dataset}:} The PPEs Dataset, provided by Roboflow, contains images of PPEs from various work environments. This dataset is used to detect and classify PPEs such as helmets, goggles, and gloves. The images are diverse, covering different workplace scenarios, ensuring a comprehensive dataset for PPE detection tasks.
    \item \textbf{CPPE-5: Medical Personal Protective Equipment Dataset 
    \citep{dagli2021cppe5}:} The CCPE-5 dataset focuses on medical personal protective equipment and includes images from hospitals and other healthcare settings. This dataset annotates medical masks, face shields, gloves, and other equipment to enhance safety detection in medical environments. It is particularly useful for improving the detection and classification of medical PPEs in clinical settings.    
    \item \textbf{Pictor-v3  \citep{nath2020deep}:} The Pictor-v3 dataset extends Pictor-v2 by adding classes for hat and vest. Relevant classes (worker, hat, and vest) are used in this study. This dataset focuses on construction site real-time PPE detection.
\end{itemize}

\begin{table}[h!]
\caption{\textbf{Directly Observable (DO)}: Attributes that are straightforward and can be directly observed from the visual features. \textbf{Situationally Observable (SO)}: Attributes that require some situational understanding but still have a significant visual component. \textbf{Inferentially Observable (IO)}: Attributes that are inferred based on context, usage, or additional knowledge about the object.}
\centering
\resizebox{\textwidth}{!}{%
\begin{tabular}{lll}
\hline
\textbf{Type}   & \textbf{Observable levels} & \textbf{Possible Values}                                                                                                               \\ \hline
Colors          & Directly Observable                      & \begin{tabular}[c]{@{}l@{}}black, dark blue, dark green, dark purple, yellow, light blue, \\light green, light purple, white, blue, brown, green, grey,\\ purple, red, \end{tabular}                                                                                                                                       \\ \hline
Materials       & Situationally Observable                     & \begin{tabular}[c]{@{}l@{}}plastic, polycarbonate, leather, rubber, latex, nitrile, fabric\end{tabular}                                                                                                                                                                                                                                                                                   \\ \hline
Functionalities & Inferentially Observable                 & \begin{tabular}[c]{@{}l@{}}shock-absorbing, impact-resistant, insulated, highly-visible, \\reflective, anti-slip, cut-resistant, puncture-resistant, \\dust-proof, fragment-proof, UV-protected, splash-proof, \\flame-retardant, chemical-protective, acid-resistant, \\alkali-resistant, face-protective, eye-protective, virus-proof, \\bacteria-proof, liquid-resistant, hair-covering, waterproof, \\contamination-preventive, stain-resistant\end{tabular} \\ \hline
\end{tabular}
}
\label{attributes}
\end{table}

As shown in Fig.\ref{data}, the proposed dataset is an attribute-based detection dataset, which covers 6 scenes, and 9 object categories, encompassing 49 attributes across image datasets. The attributes used to describe objects that are reported in table \ref{attributes}. As shown in Table \ref{attributes}, to address the varying levels of complexity in visual attribute recognition, we classify attributes into three categories based on the level of context and reasoning required to identify them accurately. Directly observable attributes are those that can be immediately identified from the visual features without the need for additional context or interpretation. Situationally observable attributes require some understanding of the scene's context but still have a significant visual component that aids in their identification. Finally, Inferentially observable attributes are those that cannot be directly seen but must be inferred through context, usage, or additional knowledge about the object. These three categories correspond to color, material, and functionality. 

\subsection{Implementation Details}
For the scene recognition module, we used BILP 2 with configuration of blip2-opt-2.7b. For the object detection model, we experiment with Yolo-world. The implementation of Yolo-world utilizes the yolov8l-worldv2 configuration in Ultralytics. Regarding the large VLM, we employ Open-CLIP. Specifically, the vision transformer encoder configuration is ViT-L/14@224px and the text encoder is OpenAI's tokenizer. We generate required safety items and visual feature attributes for each item using OpenAI ChatGPT 4o. All experiments are conducted with a single NVIDIA RTX 4090 24GB GPU.

\subsection{Compare to state-of-art methods} \label{Compare to state-of-art methods}

\begin{figure}[h!]
    \centering
    \includegraphics[width = 0.98\textwidth]{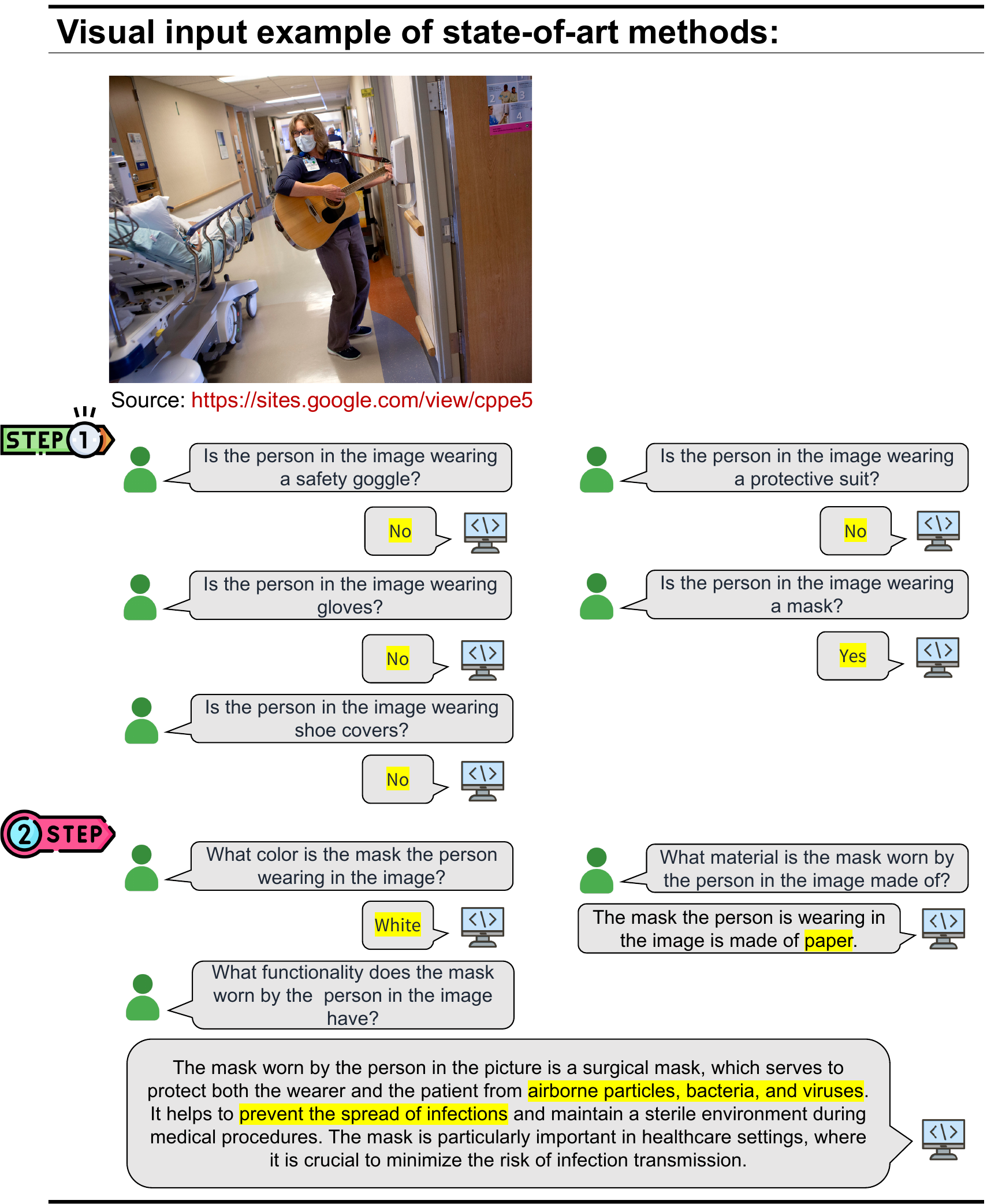}
    \caption{Example of Benchmarking LLaVA-1.6-7b for Our Safety Detection Task. Step 1 involves asking the model to identify the presence of required safety items by posing yes/no questions. Step 2 focuses on verifying specific attributes of the detected items by asking more detailed questions to ensure compliance with safety standards.}
    \label{sota}
\end{figure}

To align with our model's process of first verifying whether the required items are worn and then checking if the items have the required attributes, we chose the VQA model as our baseline. In Fig.\ref{sota}, we use LLaVA-1.6-7b to show an example of how we do our task on VQA models. Similar to our model, we divide the questions to two steps. In the first step, we utilize yes/no questions form interrogative sentences to check whether the person wearing the items. In the second step, we use wh-question form interrogative sentence to check whether the item has the required attributes. To evaluate the correctness of an answer, it is required to compare a VQA model's output with an answer. We begin with the simplest evaluation metric, called \textit{ExactMatch (EM)}. It involves preprocessing the model’s output \textit{p}, a free-format text that resembles natural language with humanlike fluency, and the corresponding answer \textit{c} by removing non-alphanumeric characters, applying lowercase conversion, and normalizing spaces. However, for \textit{EM} a prediction is considered correct only if the model's output exactly matches the correct answer. This strict criterion may not accurately evaluate the model's performance as it does not account for semantically correct but syntactically different answers. A less restrictive option is to consider a response correct if the prediction contains the true answer after preprocessing, named as \textit{Contains (Cont)} \citep{xu2023lvlm}. 
As shown in Fig.\ref{sota}, the ground truth answers are highlighted in yellow. If the output of the VQA models contains the true answer, it is considered correct. For example, in response to the question, "What functionality does the mask worn by the person in the image have?" LLaVA-1.6-7b correctly identifies that the mask can prevent airborne particles, bacteria, and viruses, and the spread of infections. These functionalities are included in the ground truth answer, so the response is marked as correct.

\begin{table}[h!]
\caption{Model details. In the prompt, ``\{\}'' is replaced by the question, QA: ``Question: \{\} Answer:'', *: LLaVA prompt, "User: <image> {} Answer:".}
\centering
\begin{tabular}{lccccl}
\hline
\multicolumn{1}{c}{\textbf{Method}} & \multicolumn{1}{l}{\textbf{\begin{tabular}[c]{@{}l@{}}Image\\ Res. (px)\end{tabular}}} & \textbf{Prompt}      & \textbf{\begin{tabular}[c]{@{}c@{}}Visual\\ Encoder\end{tabular}} & \textbf{\begin{tabular}[c]{@{}c@{}}Text\\ Encoder\end{tabular}} & \multicolumn{1}{c}{\textbf{\begin{tabular}[c]{@{}c@{}}Model\\ Size\end{tabular}}} \\ \hline
\textbf{BILPvqa}                    & \textit{480}                                                                           & \{\}                 & \textit{Vit-B/16}                                                 & \textit{BERT}                                                   & 0.361B                                                                            \\
\textbf{BILP-2 OPT}                 & \textit{224}                                                                           & QA                   & \textit{Vit-g/l EVA-CLIP}                                         & \textit{unssup. OPT}                                            & 3.745B                                                                            \\
\textbf{BILP-2 T5}                  & \textit{224}                                                                           & QA                   & \textit{Vit-g/l EVA-CLIP}                                         & \multicolumn{1}{l}{\textit{inst. FlanT5 XL}}                    & 3.942B                                                                            \\
\textbf{InstructBILP T5}            & \textit{224}                                                                           & QA                   & \multicolumn{1}{l}{\textit{Vit-g/l EVA-CLIP}}                     & \multicolumn{1}{l}{\textit{inst. FlanT5 XL}}                    & 4.023B                                                                            \\
\textbf{InstructBILP V}             & \textit{224}                                                                           & QA                   & \multicolumn{1}{l}{\textit{Vit-g/l EVA-CLIP}}                     & \textit{Vicuna-7B}                                              & 7.913B                                                                            \\
\textbf{LLaVA-1.5-7B}               & \textit{224}                                                                           & *                   & \textit{ViT-L/14 CLIP}                                            & \textit{LLaMA 7B}                                               & 6.743B                                                                            \\
\textbf{LLaVA-1.6-7B}               & \textit{224}                                                                           & *                   & \textit{ViT-L/14 CLIP}                                            & \textit{Vicuna-7B}                                              & 7.06B                                                                             \\
\textbf{Clip2Safety}       & \textit{224}       &/    & \textit{ViT-L/14 CLIP}                                            & \textit{OpenAI's tokenizer}                                                   &0.428B                                                                                   \\ 
\hline
\end{tabular}
\label{model details}
\end{table}

In Table \ref{benchmark}, we present a comparison of Clip2Safety's performance with other state-of-the-art VQA models, highlighting results for Step 1 and three observable levels in Step 2. We selected one model fine-tuned on VQAv2, BLIP \citep{li2022blip}; one multi-purpose model evaluated in a zero-shot manner, BLIP-2 \citep{li2023blip}; and two conversational models, InstructBLIP \citep{dai2023instructblipgeneralpurposevisionlanguagemodels} and LLaVA \citep{liu2024visual}. All models consist of an image encoder and a text encoder based on the transformer architecture, along with a fusion model. These models leverage pretrained unimodal encoders and train the fusion model using image-text data.
BLIP fine-tunes the encoders together with the fusion module using common losses such as image-text contrastive loss (ITC), image-text matching loss (ITM), and image-conditional language modeling loss (LM). In contrast, BLIP-2, InstructBLIP, and LLaVA keep the unimodal encoders frozen and train only the fusion module. Notably, part of the training data for InstructBLIP includes the VQAv2 dataset. This comparative analysis underscores the architectural differences and training strategies employed by each model, providing context for their performance metrics.
In Table \ref{model details}, we detail the configurations of various models. Each model's image resolution, prompt type, visual encoder, text encoder, and model size are specified. The BLIPvqa model uses a 480-pixel image resolution with a Vit-B/16 visual encoder and a BERT text encoder, totaling 0.361 billion parameters. BLIP-2 OPT employs a 224-pixel resolution, Vit-g/ EVA-CLIP visual encoder, and an unsupervised OPT text encoder, with a model size of 3.745 billion parameters. Similarly, BLIP-2 T5 uses the same visual encoder but with a FlanT5 XL text encoder, resulting in a 3.942 billion parameter model. The InstructBLIP T5 and InstructBLIP V models also utilize the Vit-g/ EVA-CLIP visual encoder but differ in their text encoders, using FlanT5 XL and Vicuna-7B, with sizes of 4.023 billion and 7.913 billion parameters, respectively. The LLaVA-1.5-7B and LLaVA-1.6-7B models feature the ViT-L/14 CLIP visual encoder and text encoders LLaMA 7B and Vicuna-7B, with sizes of 6.743 billion and 7.06 billion parameters, respectively, both at a 224-pixel resolution. Finally, the Proposed Model combines the ViT-L/14 CLIP visual encoder with OpenAI's tokenizer, resulting in a model size of 0.428 billion parameters and a 224-pixel resolution.

\begin{table}[]
\caption{Accuracy of our proposed Clip2Safety method compared to other state-of-the-art methods, evaluated step by step.}
\centering
\begin{tabular}{lccccc}
\hline
\multirow{2}{*}{\textbf{Method}} & \multirow{2}{*}{\textbf{Step 1}} & \multicolumn{3}{c}{\textbf{Step 2}}     & \multirow{2}{*}{\textbf{Mean}} \\ \cline{3-5}
                                 &                                  & \textbf{DO} & \textbf{SO} & \textbf{IO} &                                \\ \hline
\textbf{BILP vqa}                & 76.4                             & 61.6        & 53.2        & 3.2         & 48.6                           \\
\textbf{BILP-2 OPT}              & 72.0                             & 39.6        & 58.0        & 19.2        & 38.2                           \\
\textbf{BILP-2 T5}               & 78.4                             & 48.4        & 32.4        & 39.2        & 49.6                           \\
\textbf{InstructBILP T5}         & 76.8                             & 52.8        & 39.2        & 29.2        & 49.5                           \\
\textbf{InstructBILP V}          & 81.6                             & 46.4        & 60.0        & 5.2         & 48.3                           \\
\textbf{LLaVA-1.5-7b}            & 74.4                             & 54.8        & 62.8        & 82.0        & 68.5                           \\
\textbf{LLaVA-1.6-7b}            & 91.2                             & 48.4        & 62.0        & 78.8        & 70.1                           \\
\textbf{Clip2Safety}          & 76.8                             & 76.9        & 61.4        & 65.8        & 70.2                           \\
\textbf{Clip2Safety+GPT4o}    & 77.2                             & 79.7        & 66.6        & 65.8        & 72.3                           \\ \hline
\end{tabular}
\label{benchmark}
\end{table}

In Table \ref{benchmark}, we show the results of comparing our model with other VQA models.
In Step 1, LLaVA-1.5-7b and LLaVA-1.6-7b models performed notably well, achieving accuracy of 74.4\% and 91.2\%, respectively. Our Clip2Safety matched InstructBLIP T5 with an accuracy of 76.8\%, while using  GPT4o to make decision on our Clip2Safety scored 77.2\%.
In Step 2, focusing on directly observable attributes, we can see that in step 1, the powerful LLaVA is not as good as BILPvqa. Our Clip2Safety can achieve an accuracy of 76.9\%, and the accuracy of using GPT4o to make decisions can be improved by 1.8\%.
For situationally observable attributes, our model is 1.4\% lower than the best LLaVA-1.5-7b. When we use GPT4o to make decisions, the accuracy can be improved to 66.6\%. In the inferentially observable (IO) category, LLaVA-1.5-7b and LLaVA-1.6-7b excelled with 82.0\% and 78.0\%. However, in terms of average accuracy, LLaVA-1.6-7b is ahead of other VQA models. But it is still 0.1\% lower than our Clip2Safety. Clip2Safety+GPT4o can achieve the highest average accuracy of 72.3\%. And we also compare the inference time in Table \ref{inference time}. We can see that no matter step 1 or step 2, the inference time of our model is much lower than that of other VQA models.

\begin{table}[]
\caption{Inference time of our proposed Clip2Safety method compared to other state-of-the-art methods, evaluated step by step.}
\centering
\begin{tabular}{lccccc}
\hline
\multirow{2}{*}{\textbf{Method}} & \multirow{2}{*}{\textbf{Step 1}} & \multicolumn{3}{c}{\textbf{Step 2}}     & \multirow{2}{*}{\textbf{Mean}} \\ \cline{3-5}
                                 &                                  & \textbf{DO} & \textbf{SO} & \textbf{IO} &                                \\ \hline
\textbf{BILP vqa}                & 1.3                              & 1.3         & 1.3         & 1.6         & 1.4                            \\
\textbf{BILP-2 OPT}              & 7.9                              & 7.0         & 8.3         & 12.0        & 8.8                            \\
\textbf{BILP-2 T5}               & 6.1                              & 6.0         & 7.5         & 8.2         & 7.0                            \\
\textbf{InstructBILP T5}         & 13.9                             & 9.9         & 12.7        & 9.9         & 11.6                           \\
\textbf{InstructBILP V}          & 48.6                             & 49.7        & 43.2        & 46.6        & 47.0                           \\
\textbf{LLaVA-1.5-7b}            & 51.9                             & 60.5        & 107.0       & 123.2       & 85.7                           \\
\textbf{LLaVA-1.6-7b}            & 140.9                            & 144.1       & 202.3       & 237.6       & 181.2                          \\
\textbf{Clip2Safety}          & 0.4                              & 2.8         & 2.1         & 2.9         & 2.1                            \\ \hline
\end{tabular}
\label{inference time}
\end{table}

\subsection{Ablation Study}

\subsubsection{Comparison of Performance Across Different Object Detection Models}

In this section, we focus on the contribution of the different object detection models to the final accuracy. We compare the performance with the baseline, which directly detects with the original images.
To assess the impact of the object detection models, we present the accuracy for each step. The original OpenAI CLIP and OpenCLIP models were primarily designed to match images with nouns, which made it challenging for them to accurately associate each person and item with its corresponding description. As shown in Table \ref{ablation-od} (c), (e), and (g), the accuracy improves by approximately 16\%. 

Additionally, Table \ref{ablation-od} (b-d), (e-g), and (h-j) illustrate the effect of various optimizations on Clip2Safety's performance. Specifically, Table \ref{ablation-od} (b-d) utilizes Single Shot MultiBox Detector (SSD) \citep{liu2016ssd}, Table \ref{ablation-od} (e-g) employs OWL-ViT \citep{minderer2022simple}, and Table \ref{ablation-od} (h-j) uses YOLO-World \citep{cheng2024yolo}. We conduct three pairs of comparisons, namely (b) and (c), (e) and (f), and (h) and (i), to illustrate the effect of different threshold strategies (different $\delta$). In (b), (e), and (h), we utilize an average threshold of 0.6, while in (c), (f), and (i), we apply different thresholds based on g-means \citep{jain2009supervised} for each step. The accuracy improvement achieved with different $\delta$ values is 8.1\% and 0.2\% when using OWL-ViT and YOLO-World as the object detection model, respectively.

In Table \ref{ablation-od} (d), (g), and (j), we present the accuracy improvement when using GPT4o to make decision, as reported in \ref{Safety Items Detection Module}. Compared to normal settings, the LLM-decision mechanism provides 4.7\%, 11.7\%, and 2.1\% accuracy improvement when using SSD, OWL-ViT, and YOLO-world as the object detection model.

\begin{table}[]
\caption{Ablation study of different object detection models. (*OD represents Object Detection Model)}
\centering
\begin{tabular}{lccccc}
\hline
\multirow{2}{*}{\textbf{Method}}      & \multirow{2}{*}{\textbf{Step 1}} & \multicolumn{3}{c}{\textbf{Step 2}}     & \multirow{2}{*}{\textbf{Mean}} \\ \cline{3-5}
                                      &                                  & \textbf{DO} & \textbf{CO} & \textbf{IO} &                                \\ \hline
\textbf{(a) without OD}               & 60.8                             & 60.4        & 65.1        & 51.2        & 59.4                           \\
\textbf{(b) SSD}                      & 72.8                             & 73.0        & 53.1        & 54.5        & 63.4                           \\
\textbf{(c) SSD+diff $\delta$}        & 72.8                             & 62.8        & 63.1        & 53.5        & 63.1                           \\
\textbf{(d) SSD+GPT4o}                & 76.8                             & 77.6        & 64.3        & 53.5        & 68.1                           \\
\textbf{(e) OWL-ViT}                  & 72.8                             & 52.9        & 54.3        & 58.2        & 59.6                           \\
\textbf{(f) OWL-ViT+diff $\delta$}    & 72.8                             & 73.6        & 64.7        & 59.8        & 67.7                           \\
\textbf{(g) OWL-ViT+GPT4o}            & 79.2                             & 79.3        & 64.7        & 61.8        & 71.3                           \\
\textbf{(h) YOLO-World}               & 76.8                             & 76.9        & 61.4        & 65.8        & 70.2                           \\
\textbf{(i) YOLO-World+diff $\delta$} & 76.8                             & 78.8        & 63.8        & 62.2        & 70.4                           \\
\textbf{(j) YOLO-World+GPT4o}         & 77.2                             & 79.7        & 66.6        & 65.8        & 72.3                           \\ \hline
\end{tabular}
\label{ablation-od}
\end{table}

\subsubsection{Comparison of Performance Across Different Pretrained VLMs}

In Table \ref{pretrained VLMs}, we present a comprehensive comparison of various pre-trained Vision Language models, highlighting their performance across different configurations and optimization strategies. The table provides accuracy metrics for Step 1, Step 2 (DO, CO and IO), and average accuracy.

For Table \ref{pretrained VLMs} (a-f), we show the performance of using different pretrained CLIP models. Among these models, CLIP-ViT-L-patch14 achieves a mean accuracy of 70.2, which improves to 70.4 with a different threshold $\delta$. CLIP-ViT-B-patch16 shows a lower mean accuracy of 44.4\%, but applying the threshold strategy raises it to 57.1\%. The performance of CLIP-ViT-B-patch32 has a mean accuracy of 44.3\%, which improves to 63.5\% with the different threshold strategy.

For Table \ref{pretrained VLMs} (i-j), we show the performance of using different pre-trained SigLIP models, which also show notable variations in performance. SigLIP-base-patch16-384 achieves a mean accuracy of 55.6, which slightly decreases to 45.5 with the threshold strategy. SigLIP-so400m-patch14-84 and its threshold-optimized counterpart demonstrate a mean accuracy of 55.6\% and 57.1\%, respectively.

We also observe that models with larger patch sizes, such as CLIP-ViT-B-patch32, tend to benefit more from threshold adjustments due to their higher capacity and potential to capture more detailed information, which requires fine-tuning for optimal performance. Conversely, smaller patch size models, like SigLIP-base-patch16-384, exhibit more stability but less pronounced improvements, suggesting that they are less sensitive to threshold variations.

Moreover, the comparison highlights the strengths of different VLM architectures. The CLIP models generally outperform SigLIP models in terms of mean accuracy, suggesting that the architecture and pretraining strategies of CLIP models may be more effective for our tasks. However, the specific context and nature of the observable information (DO, CO, and IO) also play a crucial role in determining model performance, as seen in the varying accuracy scores across these categories.

\begin{table}[h!]
\caption{Ablation study of different VLMs. Methods include variations of CLIP and SigLIP models with different model size, patches and threshold values.}
\centering
\begin{tabular}{lccccc}
\hline
\multirow{2}{*}{\textbf{Method}}                    & \multirow{2}{*}{\textbf{Step 1}} & \multicolumn{3}{c}{\textbf{Step 2}}     & \multirow{2}{*}{\textbf{Mean}} \\ \cline{3-5}
                                                    &                                  & \textbf{DO} & \textbf{CO} & \textbf{IO} &                                \\ \hline
\textbf{(a) CLIP-ViT-L-patch14}                     & 76.8                             & 76.9        & 61.4        & 65.8        & 70.2                           \\
\textbf{(b) CLIP-ViT-L-patch14+diff $\delta$}       & 76.8                             & 78.8        & 63.8        & 62.2        & 70.4                           \\
\textbf{(c) CLIP-ViT-B-patch16}                     & 73.2                             & 26.0        & 29.5        & 49.0        & 44.4                           \\
\textbf{(d) CLIP-ViT-B-patch16+diff $\delta$}       & 61.6                             & 61.3        & 58.0        & 47.6        & 57.1                           \\
\textbf{(e) CLIP-ViT-B-patch32}                     & 73.2                             & 26.0        & 29.6        & 48.5        & 44.3                           \\
\textbf{(f) CLIP-ViT-B-patch32+diff $\delta$}       & 52.8                             & 74.1        & 64.2        & 62.9        & 63.5                           \\
\textbf{(g) SigLIP-base-patch16-384}                & 26.8                             & 74.0        & 70.4        & 51.0        & 55.6                           \\
\textbf{(h) SigLIP-base-patch16-384+diff $\delta$}  & 36.4                             & 61.3        & 48.8        & 35.3        & 45.5                           \\
\textbf{(i) SigLIP-so400m-patch14-84}               & 26.8                             & 74.0        & 70.4        & 51.0        & 55.6                           \\
\textbf{(j) SigLIP-so400m-patch14-84+diff $\delta$} & 52.8                             & 74.2        & 57.8        & 43.6        & 57.1                           \\ \hline
\end{tabular}%
\label{pretrained VLMs}
\end{table}

\subsubsection{Comparison of Performance Across Different Threshold Values}

\begin{figure}[h!]
    \centering
    \includegraphics[width = 0.98\textwidth]{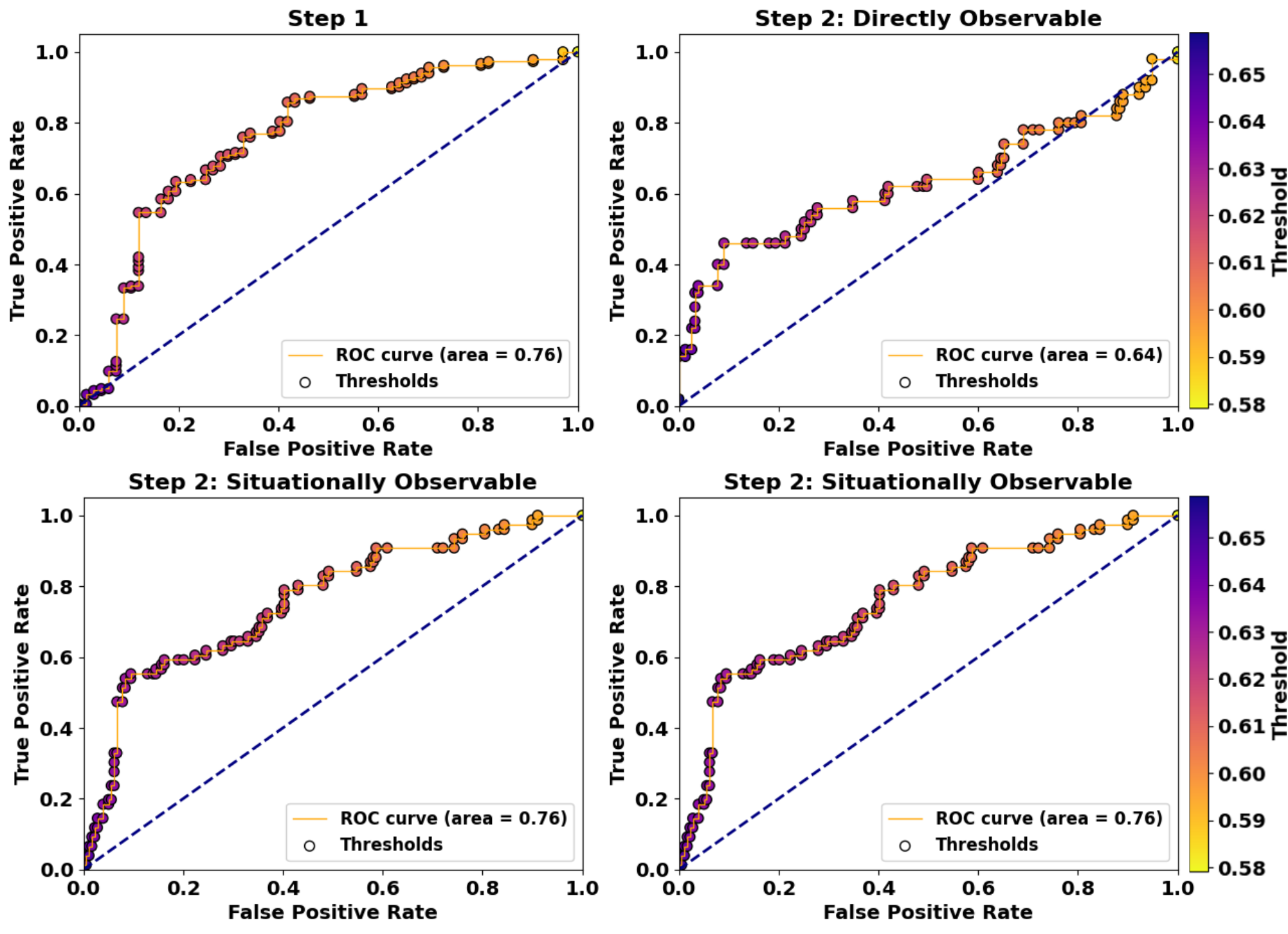}
    \caption{ROC curve analysis for different steps and threshold values. These plots show the true positive rate versus false positive rate for each step. The solid blue line represents random performance and circles indicate different threshold values, with their colors corresponding to the specific threshold level. The area under the curve values highlight the model's effectiveness in each step.}
    \label{auc}
\end{figure}

In this section, we investigate the effect of different threshold values on our model's performance. Viewing detection as a binary classification problem, Fig.\ref{auc} presents the ROC curves for various steps of our model evaluation, emphasizing the impact of varying thresholds on the true positive rate (TPR) and false positive rate (FPR) across different observable levels. We explore the range of threshold values, $\delta$, from 0.58 to 0.65, as all the similarity scores fall within this interval. 
The color gradient in each plot represents different threshold values within this range, showing their significant impact on model performance. Higher thresholds generally result in fewer false positives but may also reduce the TPR, as indicated by the tighter clustering of points near the lower end of the FPR axis. Conversely, lower thresholds increase the TPR but also raise the FPR, illustrating the trade-off between sensitivity and specificity.

The upper left plot presents the ROC curve for Step 1, which involves detecting whether a person is wearing the required item. The model achieves an area under the curve (AUC) of 0.76, indicating a strong ability to distinguish between true positives and false positives at this initial stage. The curve shows that the model maintains a high TPR while keeping the FPR relatively low, especially at moderate threshold values, which suggests that the model is effective in detecting the presence of required safety items with fewer errors.

The upper right and lower left plots show the ROC curves for the directly observable and situationally observable steps, with AUCs of 0.64 and 0.76, respectively. 
The decrease in AUC for the directly observable step compared to Step 1 indicates that as the observables become more challenging to detect, the model's performance drops. The curve for the directly observable step reveals a wider spread of points along the FPR axis, indicating increased false positives at certain thresholds. 
Despite the increased difficulty in observables compared to the directly observable step, the situationally observable step maintains a similar AUC to Step 1. This indicates the model's robustness in handling situational contexts where observables may vary depending on the scenario. The curve demonstrates that the model can still achieve a high TPR with manageable FPR at optimal thresholds, which suggests that the model is adaptable and performs well even when the context of observables is less straightforward.

The lower right plot presents the ROC curve for the inferentially observable step, which involves assessing the functionality of the item. This step has the lowest AUC of 0.55. This significant drop in performance indicates that the model struggles to accurately classify true positives and false positives when the observables require inference rather than direct detection. The curve shows a broad distribution of points, reflecting high FPRs even at higher threshold values, which illustrates that the model's sensitivity is greatly challenged in inferential scenarios.

\subsection{Visualization and Discussion}

\begin{figure}[]
    \centering
    \includegraphics[width = 0.98\textwidth]{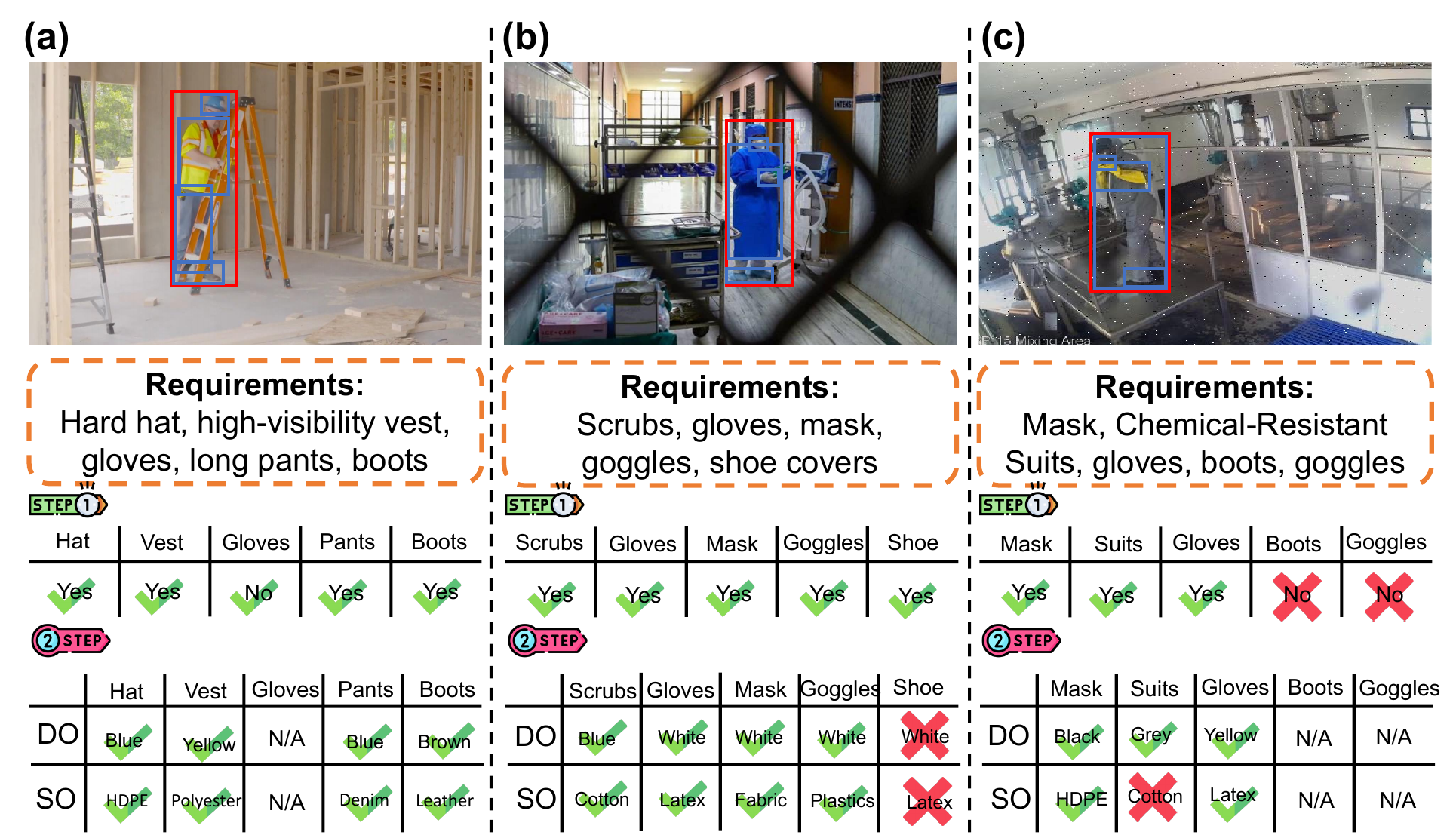}
    \caption{Visualization for detection results of the Clip2Safety. (a) Example with good performance on both step 1 and step 2. (b) Example with good performance on step 1 and bad performance on step 2. (c) Example with bad performance on step 1. (Here we only show DO and SO in step 2).}
    \label{discussion}
\end{figure}

Fig.\ref{discussion} presents our model's detection results on various image samples. Red bounding boxes label the person and blue bounding boxes label the items. Specifically, Fig.\ref{discussion} (a) showcases an example where our model performs well, with correct results in both step 1 (Safety Items Detection) and step 2 (Feature Verification). 

However, Fig.\ref{discussion} (b) highlights an instance where our model's performance is suboptimal. While the results are accurate in step 1, identifying the required safety items, the model fails in step 2, incorrectly verifying the attributes of shoes. This example illustrates a common challenge faced by our model: accurately verifying specific attributes of safety items in diverse and complex environments.

Additionally, Fig.\ref{discussion} (c) demonstrates a scenario where the model underperforms in both steps, failing to detect the required safety items and verify their attributes. This case underscores the difficulty of ensuring comprehensive safety compliance in highly variable settings, highlighting the need for further refinement and enhancement of the model to improve its robustness and accuracy across different scenarios.

\section{Conclusions} \label{Conclusion}

In this study, we introduce Clip2Safety, a novel and comprehensive framework designed to enhance safety detection across various environments. Clip2Safety efficiently leverages pre-trained knowledge and vision-language associations from the frozen CLIP model, distinguishing it from previous research efforts in this domain. 
Our approach addresses the challenge of diverse safety requirements by integrating an image captioning model to accurately interpret scene information, and a large language model to bridge the semantic gap between visual and textual data.  
Furthermore, we employ an open-vocabulary object detection model to refine the alignment of the image-text embeddings.
The efficacy of Clip2Safety is validated through empirical experiments, where it demonstrated state-of-the-art performance on our integrated dataset. Comparative analysis with existing VQA models highlights Clip2Safety's superior performance and inference efficiency. This framework not only achieves higher accuracy rates but also operates with improved computational efficiency, making it a significant advancement in the field of safety detection.

\section*{Declaration of Competing Interest}
The authors declare that they have no known competing financial interests or personal relationships that could have appeared to influence the work reported in this paper.

\section*{Acknowledgment}
This work was supported in part by the DARPA Young Faculty Award, the National Science Foundation (NSF) under Grants \#2127780, \#2319198, \#2321840, \#2312517, and \#2235472, the Semiconductor Research Corporation (SRC), the Office of Naval Research through the Young Investigator Program Award, and Grants \#N00014-21-1-2225 and \#N00014-22-1-2067. Additionally, support was provided by the Air Force Office of Scientific Research under Award \#FA9550-22-1-0253, along with generous gifts from Xilinx and Cisco.

\bibliographystyle{elsarticle-num-names} 
\bibliography{references}

\end{document}